\documentclass[conference,letterpaper]{IEEEtran} 

\IEEEoverridecommandlockouts

\usepackage[
    letterpaper, 
    left=54pt,   
    right=54pt,  
    top=72pt,    
    bottom=54pt, 
]{geometry}

\usepackage{cite}
\usepackage{amsmath,amssymb,amsfonts}
\usepackage{algorithmic}
\usepackage{tabularx}
\usepackage{graphicx}
\usepackage{textcomp}
\usepackage{xcolor}
\usepackage{url} 
\usepackage{multirow}
\usepackage{enumitem}
\usepackage{booktabs}    
\usepackage{caption}     

\def\BibTeX{{\rm B\kern-.05em{\sc i\kern-.025em b}\kern-.08em
    T\kern-.1667em\lower.7ex\hbox{E}\kern-.125emX}}

\begin{document}

\title{Empirical Performance Evaluation of Lane Keeping Assist on Modern Production Vehicles\\}

\author{
    \IEEEauthorblockN{
        \begin{tabular}{@{}c@{\hskip 1.5cm}c@{}}
            Yuhang Wang\textsuperscript{1} & Abdulaziz Alhuraish\textsuperscript{2} \\
            \small yuhangw@usf.edu & \small aalhuraish@usf.edu \\
            \smallskip
            Shuyi Wang\textsuperscript{3} & Hao Zhou\textsuperscript{4}\textsuperscript{*}\thanks{*Hao Zhou is the corresponding author (email: haozhou1@usf.edu).} \\
            \small syw@fzu.edu.cn & \small haozhou1@usf.edu \\
        \end{tabular}
    }
    \IEEEauthorblockA{
        \textsuperscript{1,2,4}\textit{Civil \& Environmental Engineering}, 
        \textit{University of South Florida}, 
        Tampa, FL, USA
    }
    \IEEEauthorblockA{
        \textsuperscript{3}\textit{Department of Transportation Engineering}, 
        \textit{Fuzhou University}, 
        Fuzhou, China
    }
}
\maketitle


\begin{abstract}

Leveraging a newly released open dataset of Lane Keeping Assist (LKA) systems from production vehicles, this paper presents the first comprehensive empirical analysis of real-world LKA performance. Our study yields three key findings:
(i) LKA failures can be systematically categorized into perception, planning, and control errors. We present representative examples of each failure mode through in-depth analysis of LKA-related CAN signals, enabling both justification of the failure mechanisms and diagnosis of when and where each module begins to degrade; (ii) LKA systems tend to follow a fixed lane-centering strategy, often resulting in outward drift that increases linearly with road curvature, whereas human drivers proactively steer slightly inward on similar curved segments; (iii) we provide the first statistical summary and distribution analysis of environmental and road conditions under LKA failures, identifying with statistical significance that faded lane markings, low pavement-laneline contrast, and sharp curvature are the most dominant individual factors, along with critical combinations that substantially increase failure likelihood. Building on these insights, we propose a theoretical model that integrates road geometry, speed limits, and LKA steering capability to inform infrastructure design. Additionally, we develop a machine learning–based model to assess roadway readiness for LKA deployment, offering practical tools for safer infrastructure planning, especially in rural areas. This work highlights key limitations of current LKA systems and supports the advancement of safer and more reliable autonomous driving technologies.

\end{abstract}

\begin{IEEEkeywords}
Lane Keeping Assist, Autonomous Driving, Empirical Analysis, Performance Evaluation, Infrastructure Design
\end{IEEEkeywords}



\section{Introduction}
\label{sec:introduction}

LKA systems, integral to ADAS, evolved from their 2003 debut on the Honda Inspire to a standardized EU mandate by 2016~\cite{honda2003inspire,unece2016r79}. Since 2022, LKA has achieved near-ubiquitous adoption, with 86.3\% of 2023 U.S. model-year vehicles equipped and all new EU vehicles have been required to have it since July 2024~\cite{katari2024driving,eu2019gsr}. As a mature technology, LKA leverages advanced camera-based lane detection and control algorithms, with ongoing refinements to address complex scenarios~\cite{ziebinski2017review}.

LKA markedly improves traffic safety by mitigating lane-departure crashes, which account for approximately 37\% of U.S. crashes~\cite{singh2015critical}. Studies on lane-departure warning (LDW), a precursor to LKA, report 11--21\% reductions in relevant crashes~\cite{cicchino2018effects}. The European Commission projects that LKA, alongside other safety systems, will prevent over 25,000 fatalities and 140,000 serious injuries by 2038~\cite{ec2024lives}. Additionally, LKA fosters smoother traffic flow by maintaining consistent lane discipline, minimizing disruptive maneuvers~\cite{ziebinski2017review}.

Despite these safety gains, LKA systems exhibit persistent operational deficiencies. Inaccurate lane detection under rain, complex merge dynamics, high-curvature roadways, or adverse weather undermines reliability, driven by misperception of lane boundaries, uncertain trajectory planning, and actuator limitations~\cite{singh2015critical,ziebinski2017review,wei2023state}. However, due to the lack of comprehensive understanding of the failure of the LKA system, some drivers may be gullible to believe in the over-advertising of some car makes and slacken off their driving attention after switching on the LKA, thus posing a safety hazard. Some research point out when automating lane-keeping, LKA reduces driver cognitive demand, particularly in monotonous highway settings, enhancing reaction times~\cite{deguzman2023factors, huang2023beyond}.

Existing research struggles to address these challenges comprehensively, constrained by several limitations. Many studies rely on proprietary or limited datasets, restricting their generalizability to varied driving contexts~\cite{ziebinski2017review, sang2025improved}. Investigations often focus on single vehicle models, failing to capture the diversity of LKA implementations across manufacturers~\cite{wei2023state, nidamanuri2021progressive}. Controlled test environments, such as test tracks or simulators, dominate performance evaluations, overlooking the complexity of real-world road and weather conditions~\cite{iqbal2023metamorphic}. Similarly, reliance on synthetic perturbations in simulations oversimplifies environmental variables, limiting applicability to edge cases like degraded markings or extreme curvature~\cite{rotili2024modeling}. Furthermore, few studies explore multi-sensor fusion strategies, which are critical for robust lane detection in adverse conditions~\cite{alrousan2021multi}. The interplay between LKA performance and road infrastructure, such as lane marking quality or curvature design, remains underexplored, despite its relevance to traffic safety~\cite{jimenez2016advanced}. Additionally, vulnerabilities in edge cases, such as heavy rain or low-visibility scenarios, are rarely addressed systematically~\cite{mehta2023securing}. These gaps underscore the need for a holistic, data-driven evaluation of LKA performance in real-world settings.

In summary, this paper makes the following key contributions:
\begin{enumerate}
    \item We conduct the first comprehensive empirical study of real-world LKA failures, systematically categorizing them into perception, planning, and control errors, supported by detailed analysis of LKA-related CAN signals to identify when and where failures emerge.
    
    \item We quantitatively evaluate LKA lane-keeping performance on curved roads, showing that current systems exhibit significantly greater deviation and reduced adaptability compared to human drivers, particularly under high-curvature conditions.
    
    \item We present the first statistical summary and distribution analysis of environmental and roadway factors associated with LKA failures, identifying dominant contributors such as faded lane markings, low contrast, sharp curvature, and their critical combinations.
    
    \item To address infrastructure compatibility with LKA limitations, we develop two modeling tools: (i) a dynamic model linking vehicle speed, curvature, and LKA steering constraints to support curvature-aware geometric design and speed posting, and (ii) a machine learning–based predictive model that distills real-world failure patterns to assess infrastructure readiness and risk for LKA deployment.
\end{enumerate}

\section{Background and Dataset}

\subsection{Related Works}

LKA system, a cornerstone of ADAS, integrates camera-based lane detection, signal processing, and lateral control to ensure vehicle alignment within marked lanes. The evolution of ADAS reveals a shift from rule-based edge detectors to learning-based lane segmentation and multi-sensor fusion, enhancing steering precision and passenger comfort~\cite{ziebinski2017review,nidamanuri2021progressive,kim2023robust}. While LKA exhibits marked robustness on nominal highways, it falters with faded or occluded markings, complex curvature, or low-visibility conditions, often confounding shadows or road edges~\cite{wei2023state,nidamanuri2021progressive,mehta2023securing}. Despite these advancements, such limitations necessitate further improvements in perception robustness, fault-tolerant planning, and adaptive torque allocation to broaden LKA’s operational design domain.

Impact assessments demonstrate significant crash reductions for LKA and its precursor, Lane Departure Warning (LDW): lane-departure crashes constitute approximately 37\% of U.S. road accidents~\cite{singh2015critical}, with LDW-equipped vehicles exhibiting 11--21\% fewer police-reported incidents~\cite{cicchino2018effects}. ADAS reviews affirm these benefits, forecasting significant societal gains with wider adoption~\cite{iihs2021benefits,aleksa2024impact}. However, these insights are constrained by limited datasets and methodological shortcomings. Many studies rely on short-term field tests or simulator trials, omitting adverse weather, rural road geometries, or long-term driver adaptation. Consumer evaluations prioritize idealized conditions, often inflating real-world performance estimates~\cite{wei2023state}. A pressing gap remains in comprehensive, open-access longitudinal analyses capable of capturing rare but safety-critical LKA failure modes across diverse operational contexts.

Roadway attributes profoundly impact camera-based ADAS performance. Empirical studies reveal that high-contrast, retro-reflective lane markings markedly enhance perception reliability, whereas faded or inconsistent markings trigger recurrent LKA disengagements and unintended lane departures~\cite{jimenez2016advanced,wei2023state}. Beyond markings, road geometry poses challenges: small-radius curvature and ambiguous merge zones often exceed actuation limits, causing torque saturation and loss of control stability~\cite{mehta2023securing}. To address these challenges, researchers advocate enhanced marking standards, standardized signage, and connected roadside units broadcasting lane geometry~\cite{aleksa2024impact}. Yet, regulatory frameworks lag: UN R79 and EU General Safety Regulation prioritize vehicle functionality over infrastructure standards~\cite{unece2016r79,eu2019gsr}. This disconnect between automation and infrastructure poses a pressing barrier, necessitating coordinated standards to align roadway design with autonomous fleets.

\subsection{The OpenLKA Dataset}

Our work on analyzing LKA systems in the real world is based on the OpenLKA dataset, a comprehensive resource for evaluating LKA systems. This dataset comprises driving data collected from two primary sources. One part consists of over 200 hours of data specifically gathered in Florida starting in the summer of 2024, designed to test LKA functionality across a variety of prevalent car models. In particular, data collection in this part focused on LKA usage, with drivers encouraged to utilize the feature as much as possible without violating traffic laws or creating hazardous situations, resulting in a rich collection of relevant data. The second part of the OpenLKA dataset originates from the Comma Community open-source automated driving platform, contributed by over 80 drivers spanning more than 30 regions worldwide, accumulating over 80 hours of everyday driving data. This segment includes instances where community members used ADAS features, as well as their regular human driving data. In total, OpenLKA encompasses over 60 distinct car models, representing a significant portion of the North American and European markets, and covers a wide array of driving scenarios, including mountainous terrain, heavy snow, urban environments, and rural roads. OpenLKA contains detailed vehicle control data which is precisely aligned with the corresponding video data. Furthermore, the dataset employs Vision Language Models (VLMs) with Chain-of-Thought (CoT) prompting techniques to extract valuable labels, such as weather conditions and traffic information, which are not directly obtainable from CAN bus data or video footage. The extensive scale and comprehensive nature of the OpenLKA dataset make it an ideal resource for in-depth analysis of LKA system performance.

\subsection{Curation of LKA Failure Dataset}

To facilitate the analysis of LKA performance, we performed several preprocessing steps on the OpenLKA dataset. We utilized the output distance to left and right laneline from the Openpilot's end-to-end model to determine the vehicle's position in the lane. Subsequently, we get the \textbf{OpenLKA-Failure} dataset based on the position and CAN to identify instances where the vehicle exhibited significant lane deviations, exceeding 0.25 meters, or the CAN bus data indicated an LKA disengagement event. To provide a basis for comparison, we create \textbf{OpenLKA-Normal} dataset, which is randomly sampled a comparable amount of data representing normal LKA operation, where the system maintained the vehicle centering the lane without disengagement. This approach allowed us to analyze both failure cases and typical operational scenarios.

\subsection{Performance Metrics and Evaluation Methods}
\label{subsec:analysis_metrics}

\begin{figure}[t]           
  \centering
  \includegraphics[width=0.3\textwidth]{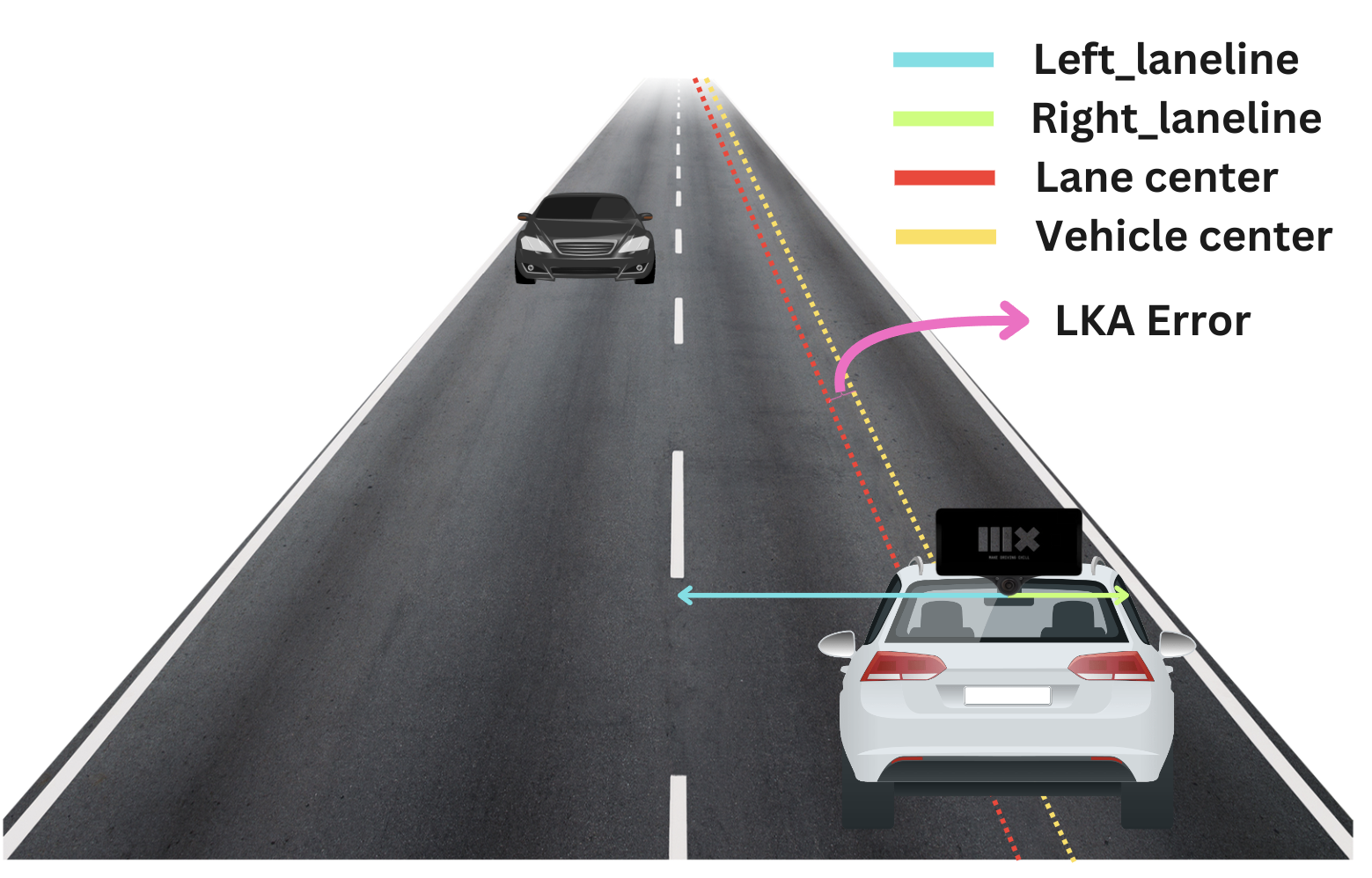} 
  \caption{Lane Position Evaluation Metrics}
  \label{fig:lka_error}
  \vspace{-0.5em}            
\end{figure}

To quantitatively assess LKA performance and identify failure modes within the OpenLKA dataset, we defined and utilized several key metrics derived from synchronized CAN logs, video data, and Openpilot outputs:

\begin{itemize}
    \item \textbf{LKA Disengagement Status:} Sourced directly from the vehicle's CAN, this binary metric indicates whether the LKA system is actively engaged or has disengaged. A disengagement (status 0) is considered an LKA system failure or inability to operate under the current condition.

    \item \textbf{Lane Deviation (Lane Centering Error):} This metric quantifies the vehicle's lateral deviation from the lane center. It is calculated using the vehicle's distance to the left and right lanelines, obtained from Openpilot:
    \begin{equation}
    \text{Lane Deviation} = \frac{d_{\text{left}} + d_{\text{right}}}{2}
    \end{equation}
    The calculation is illustrated in Fig.~\ref{fig:lka_error}. We classify LKA performance based on this error: an absolute deviation exceeding 0.25 meters is considered an anomaly (significant deviation). An absolute deviation greater than 0.65 meters, or any LKA disengagement event, is classified as a critical LKA failure. 

    \item \textbf{Laneline Detection Result:} We analyzed lane line detection performance using two distinct sources:
        \begin{itemize}
            \item \textit{Vehicle System (CAN):} A discrete 'Lane Detection Level' obtained from the CAN, ranging from 0 to 3. Level 0 signifies complete non-detection; Level 1 indicates clear detection; Level 2 represents ambiguous or blurred detection; and Level 3 denotes clear detection of a special lane line type (e.g., dashed, yellow).
            \item \textit{Openpilot Model:} A continuous probability score (0\% to 100\%) from Openpilot's end-to-end vision model, representing the confidence in lane line detection. We interpret this as: $\geq 90\%$ = normal detection; $80\%$ to $<90\%$ = ambiguous detection; $<80\%$ = problematic detection. Given that Openpilot utilizes SOTA model architectures, its detection results are generally considered more robust than those from the native vehicle systems, which often rely on older algorithms (typically pre-2022).
        \end{itemize}

    \item \textbf{Road Curvature:} We directly used the road curvature value estimated by Openpilot.
\end{itemize}

These metrics, along with VLM-annotated contextual labels (weather, lighting, etc.) converted into structured features, provide a strong foundation for a layered analysis of LKA performance.

\section{Empirical LKA Performance Analysis}

\subsection*{LKA Performance under Normal Conditions}
\label{subsec:normal_conditions}

\begin{figure}[t]
  \centering
  \includegraphics[width=0.48\textwidth]{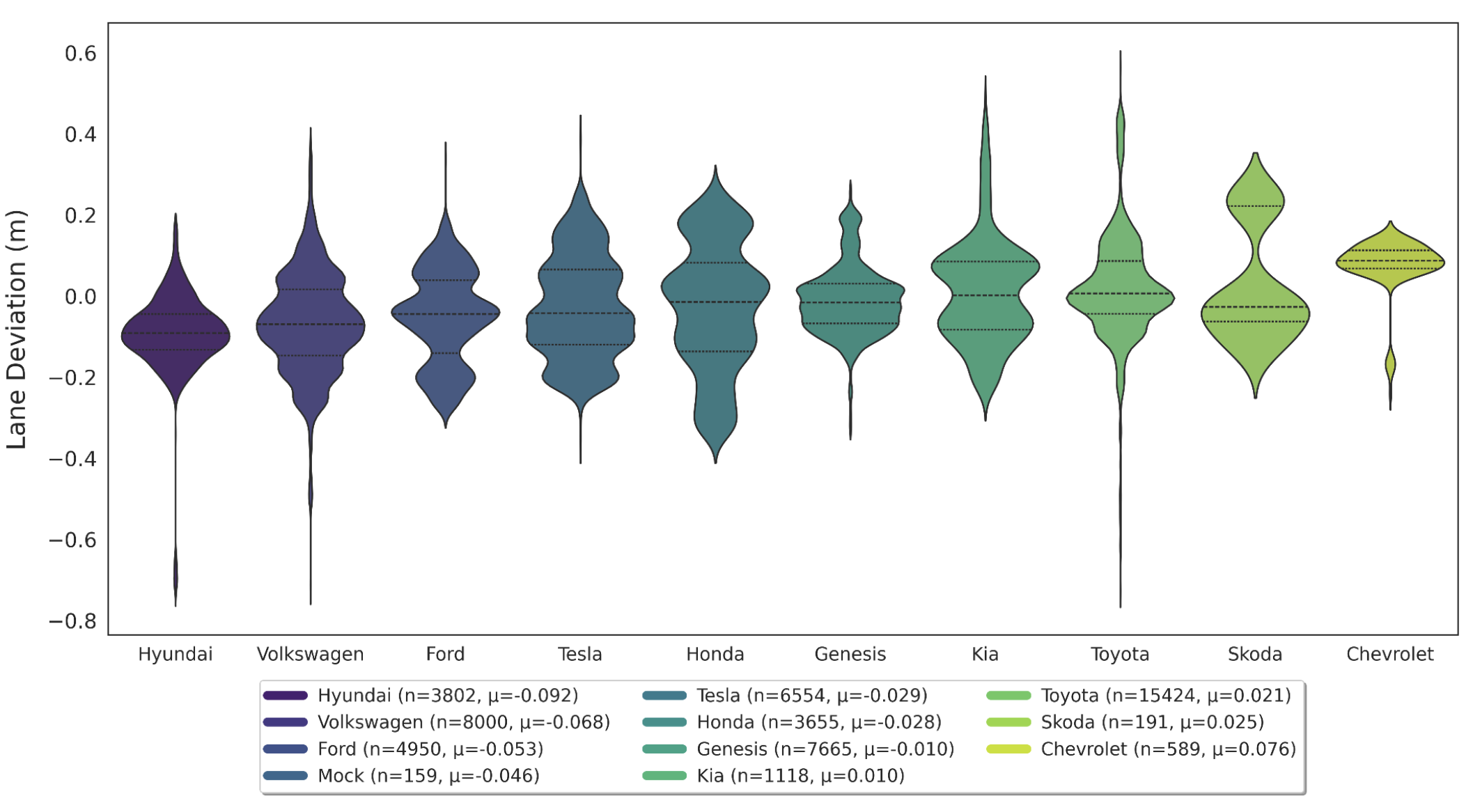}
  \caption{Lane Deviation: Curvature is less than 0.001, and the Openpilot's lane detection probability is greater than 0.9. The road curvature is low, the lane detection effect is acceptable, and the significance value reflects the performance under normal conditions to a certain extent.}
  \label{fig:normal_deviation}
\end{figure}

\begin{figure}[t]
  \centering
  \includegraphics[width=0.48\textwidth]{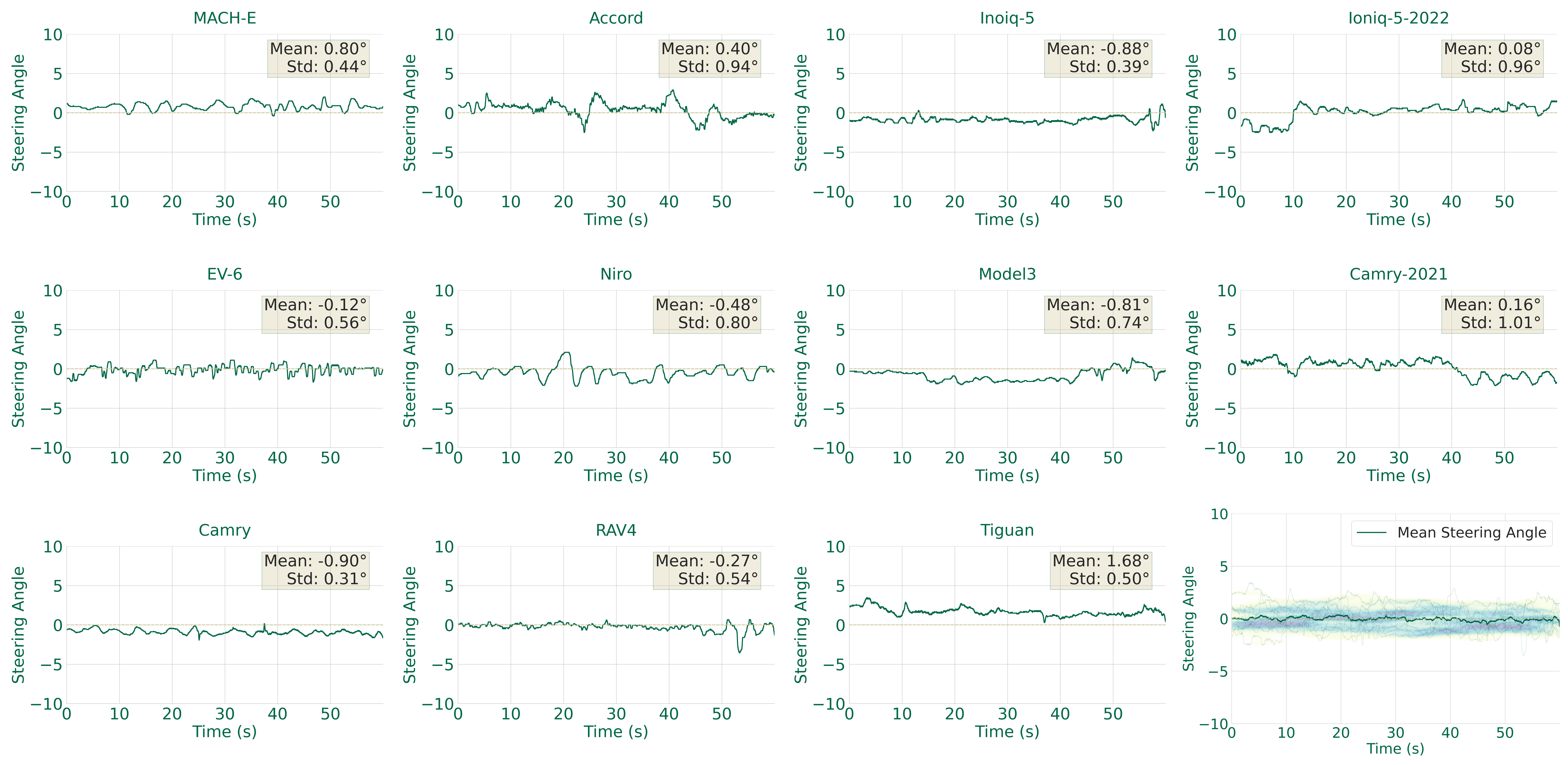}
  \caption{Steering stability under scenarios whose curvature is less than 0.0005 and the lane detection results are greater than 0.95.}
  \label{fig:normal_steering}
\end{figure}

For the purposes of this study, \textit{Normal Conditions} are defined as scenarios characterized by clear visibility, flat road surfaces, low road curvature, and clearly visible lane markings. These conditions represent ideal operating environments for LKA systems, where external factors such as weather or road geometry pose minimal challenges. To evaluate LKA performance under these conditions, relevant scenarios were sampled from the OpenLKA-Normal and OpenLKA-Failure dataset, ensuring a representative subset of driving data across diverse vehicle models and geographic regions.

Analysis revealed that, under normal conditions, the lane deviation, consistently remained within 0.2 meters (Fig.~\ref{fig:normal_deviation}). This indicates a high degree of lane-keeping accuracy, with the vehicle maintaining close alignment to the lane centerline. Furthermore, the steering angle, a key indicator of control stability, exhibited minimal variation, reflecting smooth and predictable steering behavior without abrupt or significant corrections. These findings, illustrated in Fig.~\ref{fig:normal_steering}, demonstrate that LKA systems perform robustly in normal conditions, effectively maintaining lane discipline and operational stability~\cite{kim2023robust}.

\subsection*{Categorization and Diagnosis of Common LKA Failures}

\begin{figure}[b]
  \centering
  \includegraphics[width=0.48\textwidth]{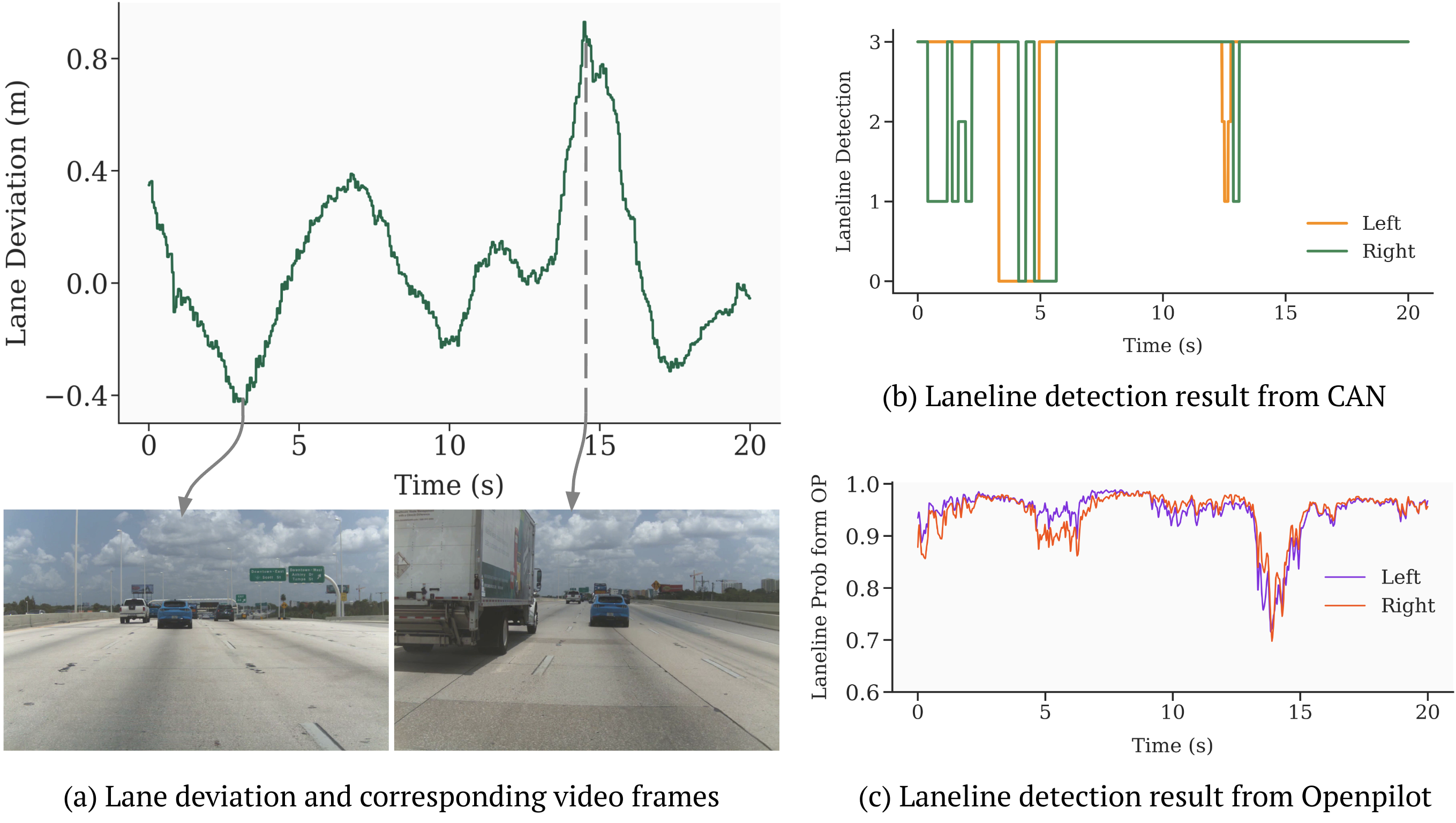}
  \caption{Low pavement-laneline contrast (white pavement white laneline) causes lane detection failure}
  \label{fig:lowcontrast}
\end{figure}

Adverse conditions, such as sharp road curves, heavy rain, glaring sunlight, degraded lane markings, and their combinations, pose significant challenges to LKA systems. These scenarios, extensively tested in the OpenLKA dataset and this subsection will analyze these limitations through the lenses of perception, planning, and control, leveraging empirical findings from the dataset to quantify performance degradation and identify safety risks.

\subsubsection{Perception}
\label{subsubsec:perception}

LKA systems rely heavily on accurate lane detection to maintain vehicle positioning, yet their perception modules exhibit pronounced vulnerabilities in adverse conditions. While current lane detection algorithms demonstrate considerable robustness in detecting lane lines even under conditions of partial occlusion or absence, as exemplified by state-of-the-art methods achieving high accuracy, $97+\%$ on benchmark datasets like Tusimple~\cite{CLRNet2022CVPR}, real-world driving scenarios present a vast array of long-tail events where multiple challenging factors often occur concurrently, posing significant difficulties. 

\begin{figure}[h!]
  \centering
  \includegraphics[width=0.48\textwidth]{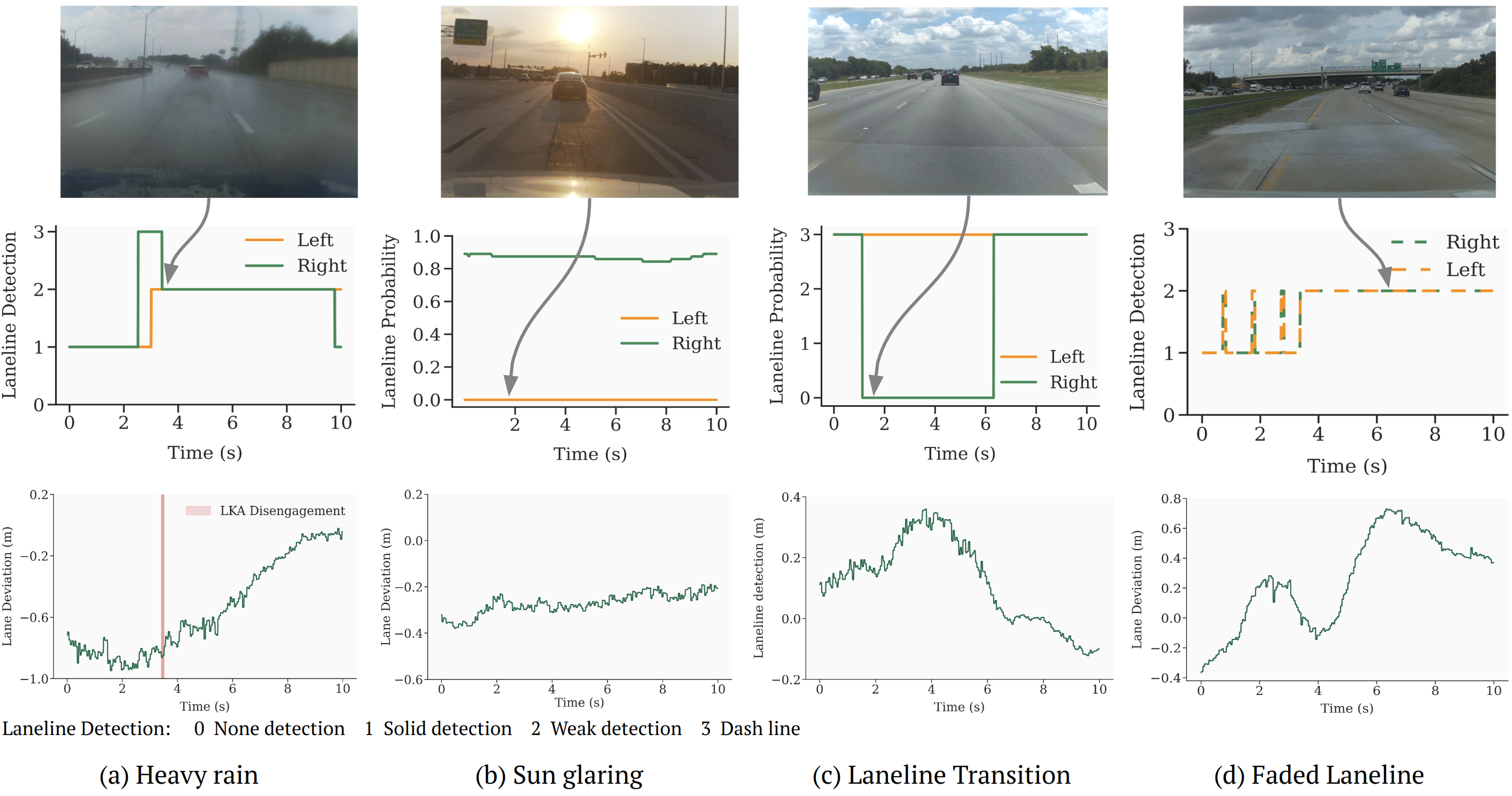}
  \caption{Common Cases in Laneline Detection Failure}
  \label{fig:Percepsion_cases}
\end{figure}

In the case of low pavement laneline contrast, the pavement and laneline are too close due to color and reflectivity, which will increase the difficulty of edge segmentation of road marking detection and cause detection problems. Fig.~\ref{fig:lowcontrast} is an example from OpenLKA-Failure. The vehicle is driving on a white road with white markings. (a) shows two serious lane deviations. The first is a 0.4 meters deviation to the right, and the second is a 0.8 meters deviation to the left, almost deviating from the edge of the lane. At the same time, a large truck passed by on the left. If the driver had not taken over immediately, this would have caused serious traffic accident. In addition, according to our findings, faded lanelines caused by long-term use and wear of the road will reduce the confidence of road detection. Rain splashed by vehicles on the road on rainy days and snow covering lane lines on snowy days will block lanelines and cause challenges in lane line detection. Overexposure of lane lines due to glaring direct sunlight will also increase the difficulty of edge detection. In addition, in some traffic scenarios, such as ramps and merges, lane lines will disappear. At this time, the detection algorithm of some vehicles can continue to detect the correct lane line position, but some cars cannot detect the missing lane line or mistakenly detect another lane line, causing the vehicle to cause a large deviation. These common examples are shown in Fig.~\ref{fig:Percepsion_cases}.

\subsubsection{Planning}
\label{subsubsec:planning}

\begin{figure}[h!]
  \centering
  \includegraphics[width=0.48\textwidth]{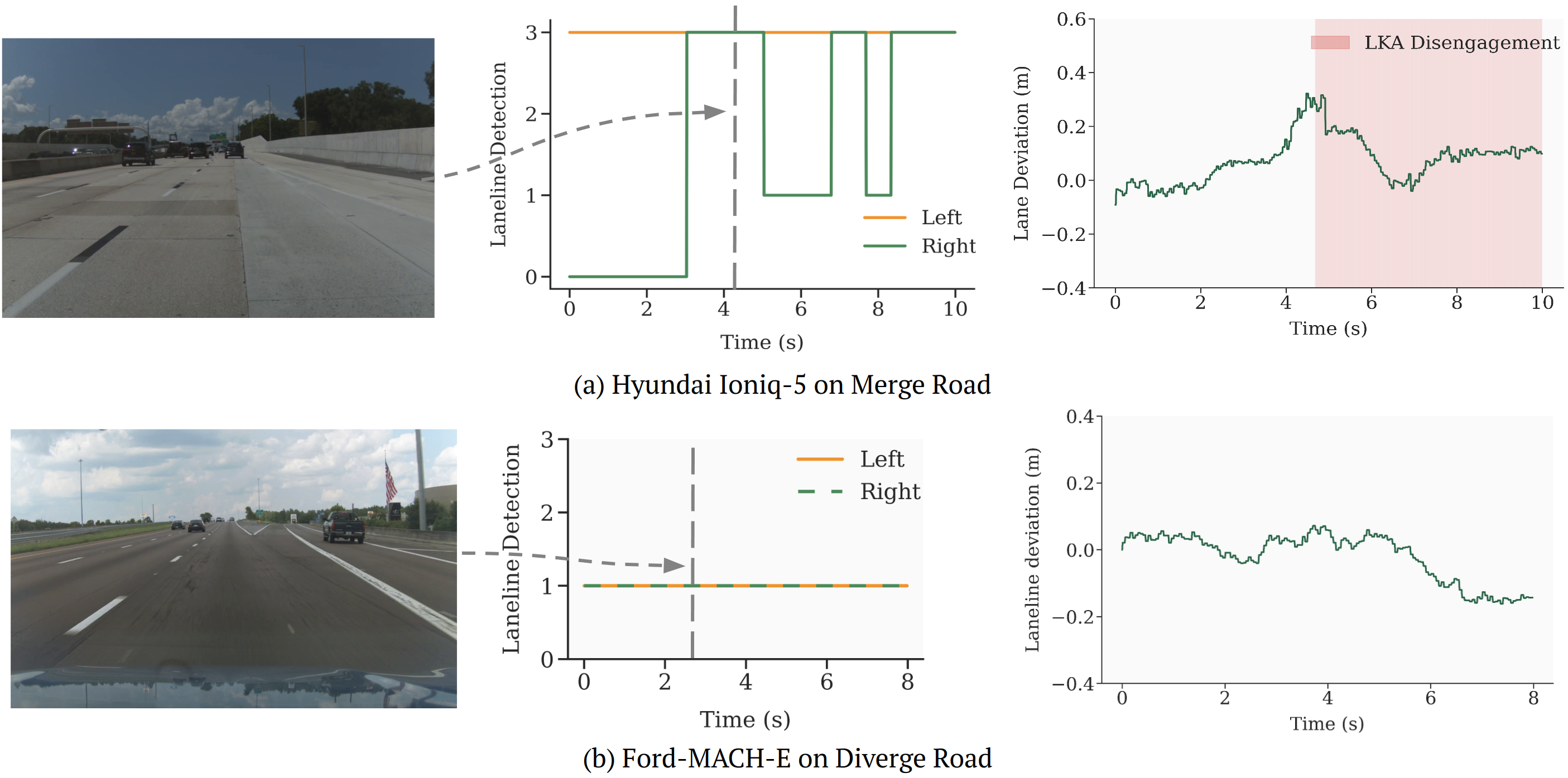}
  \caption{Hyundai Ioniq5 and Ford MACH-E's Performace in Merge and Diverge Road Segments}
  \label{fig:planning}
\end{figure}

We define planning as the part of the LKA system that provides planning information for controlling vehicle motion after obtaining the Lane detection result. The LKA system not only requires an accurate lane detection system, but it also relies heavily on the decision-making planning part. When the lane detection is incomplete or the actual lane detection is incorrect, the planning part of some vehicles can correct the incorrect lane detection results or correctly plan the vehicle's driving based on only the lane detection results with higher confidence. For example, Fig.~\ref{fig:planning} provides such an example. Figure (a) shows the performance of the Hyundai Ioniq 5 in a merge section with clear lane detection. Due to the change in the right side, the vehicle deviates to the left by 0.4m, and even LKA disengagement occurs in the second half. Similarly, as shown in Figure (b), although the Ford MACH-E is located in a diverge section and lane detection is prone to problems, the entire driving remains in the center of the lane. We speculate that the planning part of the Ford MACHE has a special design that can actively follow the unchanged left lane on sections where the road markings change. , thus stably keeping the vehicle within the lane.

\subsubsection{Control}
\label{subsubsec:control}

\begin{figure}[h!]
  \centering
  \includegraphics[width=0.48\textwidth]{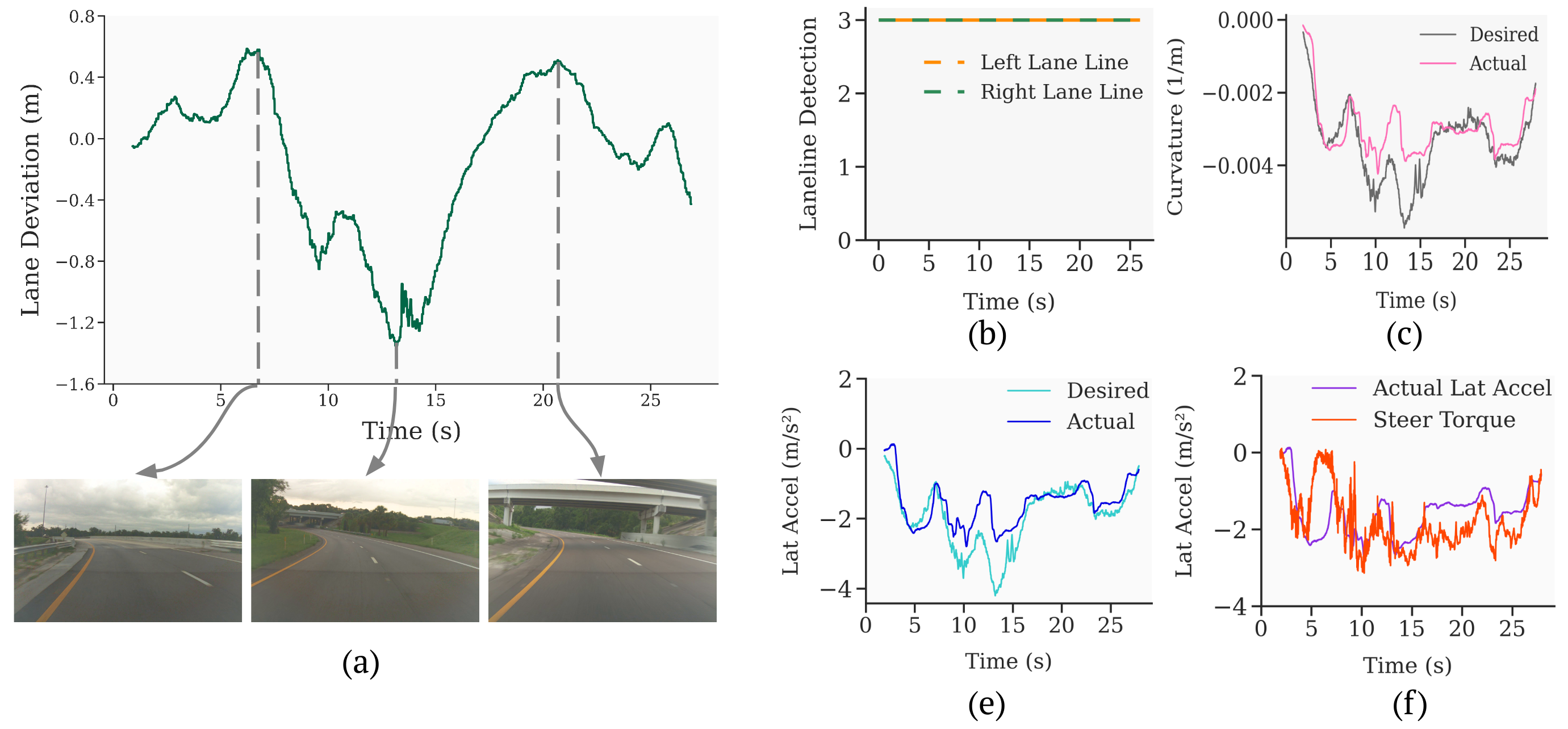}
  \caption{Hyundai Ioniq5's Lane Deviation in Sharp Curve.}
  \label{fig:sharpcurve}
\end{figure}

We have deeply studied the Lane keeping related variables of the CAN system of the vehicle computer system and found that this may be due to the Torque limit or steering angle rate limit used to ensure safety, which causes the vehicle to be unable to provide sufficient lateral acceleration when dealing with large curvature curves, resulting in the vehicle being unable to stay in the center of the lane. LKA driving is a process of gradually increasing the steering angle, which causes the lateral acceleration of the vehicle to be unable to keep up with the gradual increase in the curvature of the road, causing the vehicle to deviate from the center of the lane line or even deviate from the current lane. As shown in Fig.~\ref{fig:sharpcurve}, Hyundai Ioniq 5 faces a sharp left. When the vehicle was on a curve road, the increasing lateral acceleration could not match the curvature of the road, resulting in a deviation of 1.2 meters to the right.

\subsubsection{Multiple Factors}
\label{subsubsec:multifactors}

\begin{figure}[h!]
  \centering
  \includegraphics[width=0.48\textwidth]{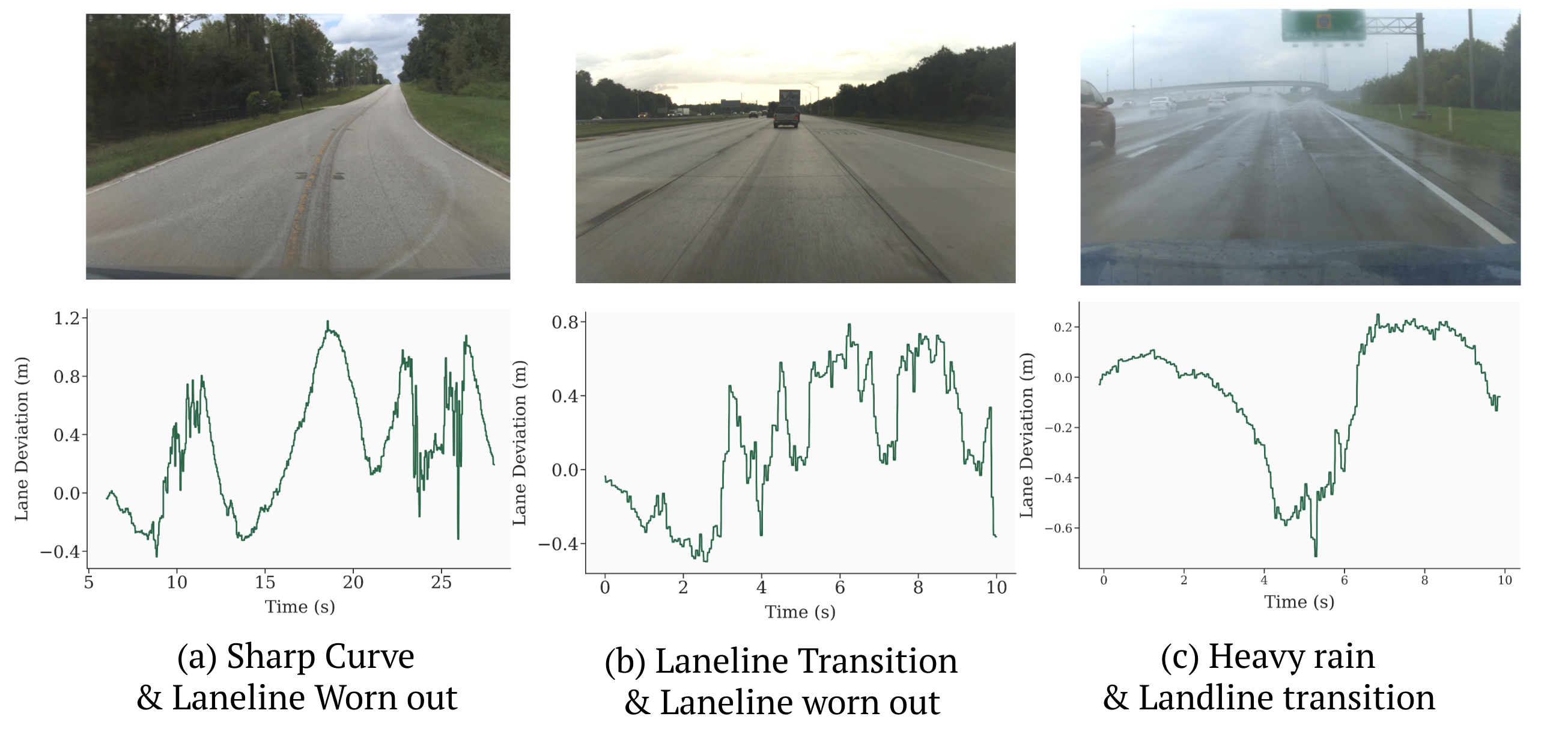}
  \caption{Multiple Factors Leading to Larger Lane Deviation}
  \label{fig:multifacs}
\end{figure}

As shown in Fig.~\ref{fig:multifacs}, we selected typical scenarios from OpenLKA-Failure. When multiple factors are superimposed and more significant, the vehicle often deviates by more than 0.8 meters or more, and the driver needs to take over immediately. If the driver is not careful, he will drive into other lanes or even risk colliding with the vehicle in the next lane.

\subsection*{LKA Performance on Curves}
\label{subsubsec:curvature}

\begin{figure}[h!]
  \centering
  \includegraphics[width=0.48\textwidth]{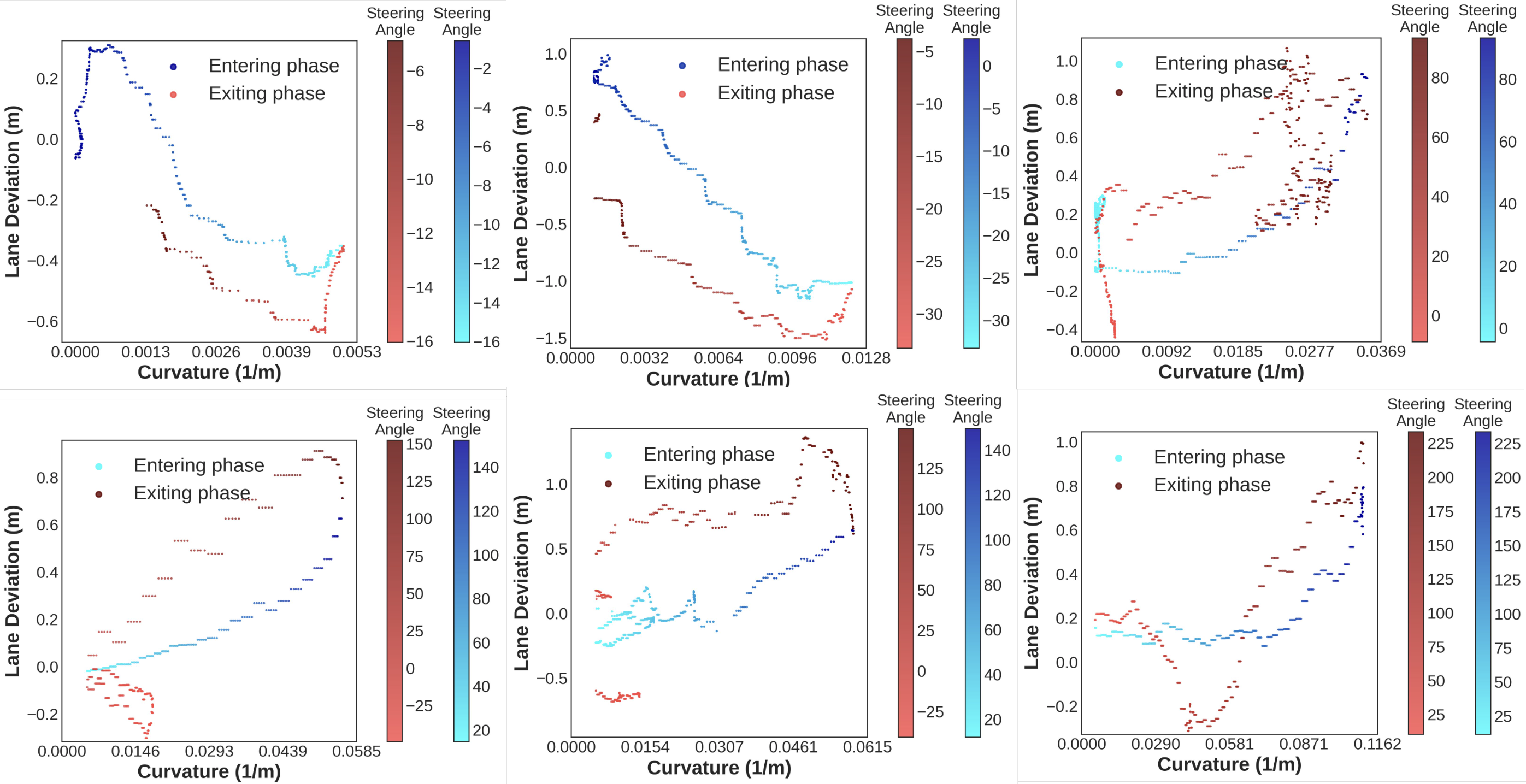}
  \caption{LKA Performance on Different Road Curvature.}
  \label{fig:curvature_deviation}
\end{figure}

To qualify the relationship between lane deviation and road curvature, we first obtain the whole process of the vehicle entering and exiting the curve, and then intercept the data points located in the first 40\% of the absolute value of the curvature, which indicates that the vehicle is located around the center of the curvature of the curve at this time. Then we regress these data points. The results (Fig.~\ref{fig:curvature_deviation}) show that the offset distance of the vehicle compared to the center of the lane has a strong linear relationship with the curvature of the curve, and as the absolute value of the curve increases, the deviation of the vehicle under the LKA system will increase, and it will deviate to the other side of the curve direction. See Fig.~\ref{fig:curvature_deviation}, This relationship can be quantified as 

\begin{equation}
\hat{d}_{\text{lane}}\;(\mathrm{m}) \;=\; -\,8.327 \;\kappa \;+\; 0.214, \qquad R^{2}=0.673
\label{eq:curvature_lane_deviation}
\end{equation}

The negative coefficient of the curvature term (-8.327) indicates that as the absolute value of the road curvature increases, the vehicle tends to deviate further from the lane center in the opposite direction of the curve. The constant term (0.214) represents a normal offset.

\begin{figure}[h!]
  \centering
  \includegraphics[width=0.32\textwidth]{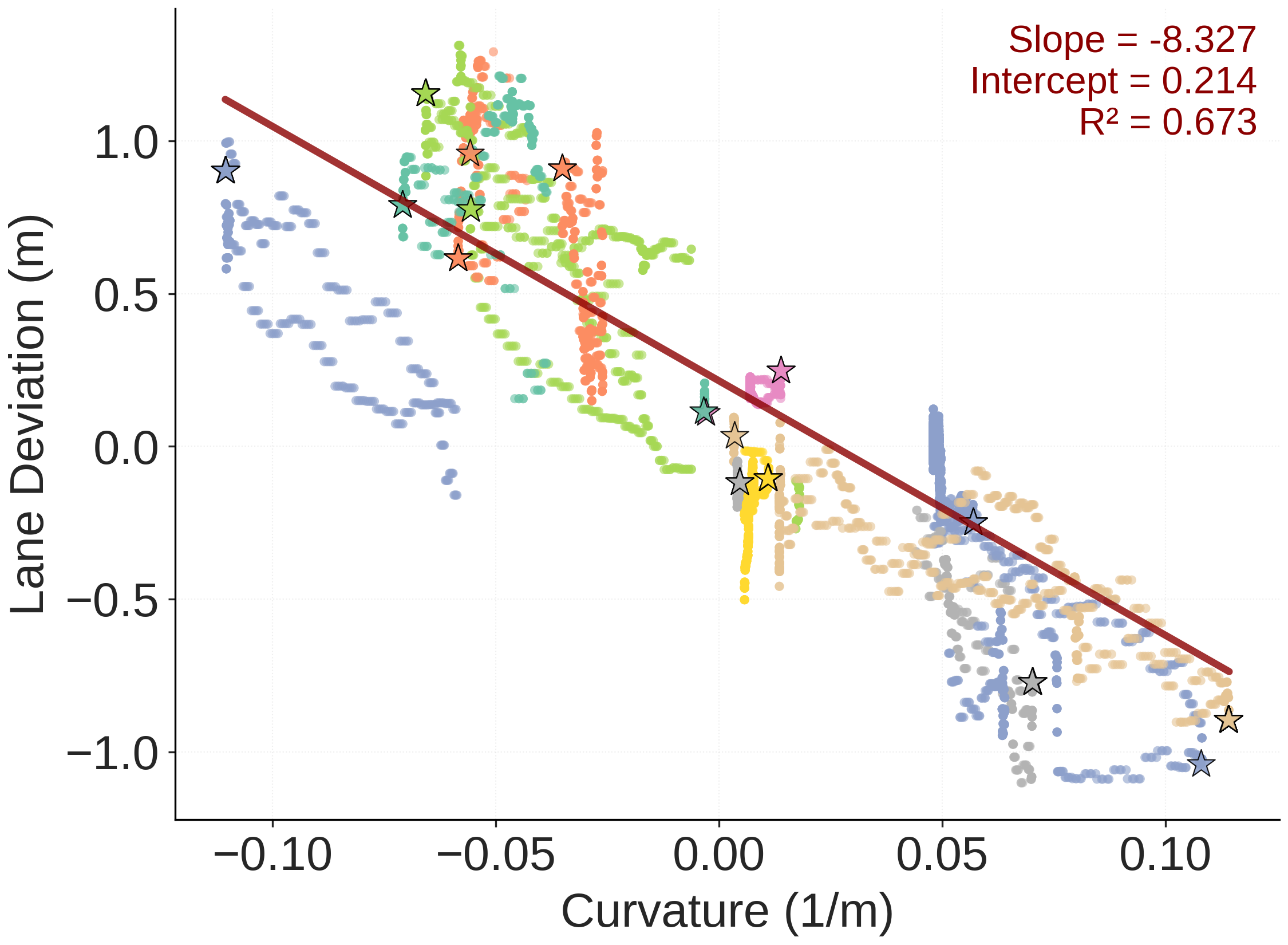}
  \caption{The relationship between Lane Deviation and Road Curvature.}
  \label{fig:fit_lanedeviation}
\end{figure}

\subsection*{Comparison of the flexibility of LKA and human driving}
\label{subsubsec:flexibility}

\begin{figure}[h!]
  \centering
  \includegraphics[width=0.48\textwidth]{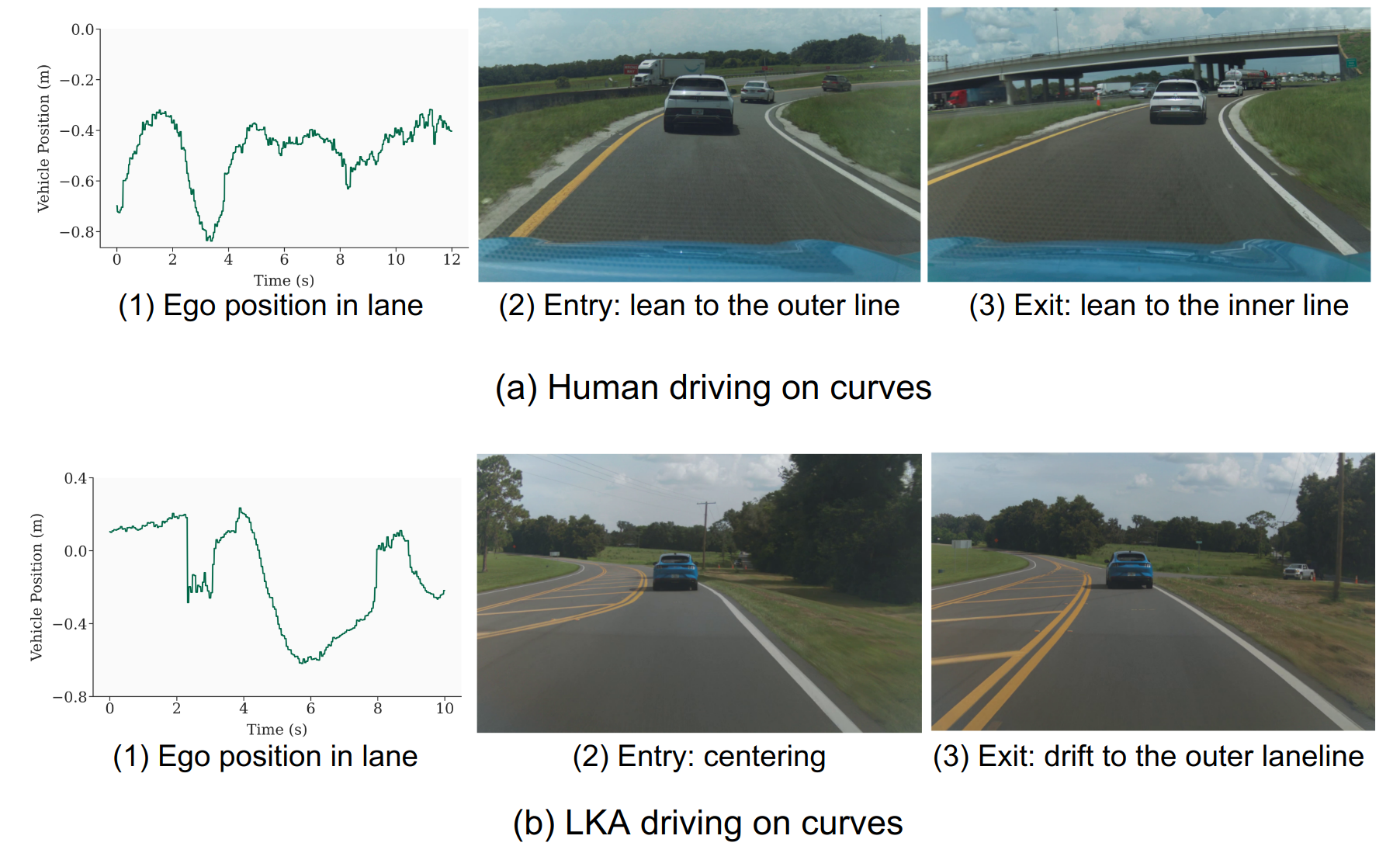}
  \caption{Comparisons of LKA and Human Driving on curves.}
  \label{fig:flexibilityoncurve}
\end{figure}

Comparing with the human driving in OpenLKA, we also found that human drivers with good driving skills tend to use corner apexing when passing through curves, that is, when entering the curve, they tend to lean to the inside of the curve to ensure that when exiting the curve, even if there is a deviation, they will not deviate a large distance or deviate from the current lane, as shown in the upper part (a) of Fig.~\ref{fig:flexibilityoncurve}. However, LKA follows rigid lateral control, which gradually adjusts the steering angle as the curve turns, can easily cause the vehicle to deviate largely in the lane. Therefore, for the LKA system that lacks flexibility, we should design a more flexible LKA system that is more in line with the style of human drivers to ensure driving safety. 

\begin{figure}[h!]
  \centering
  \includegraphics[width=0.48\textwidth]{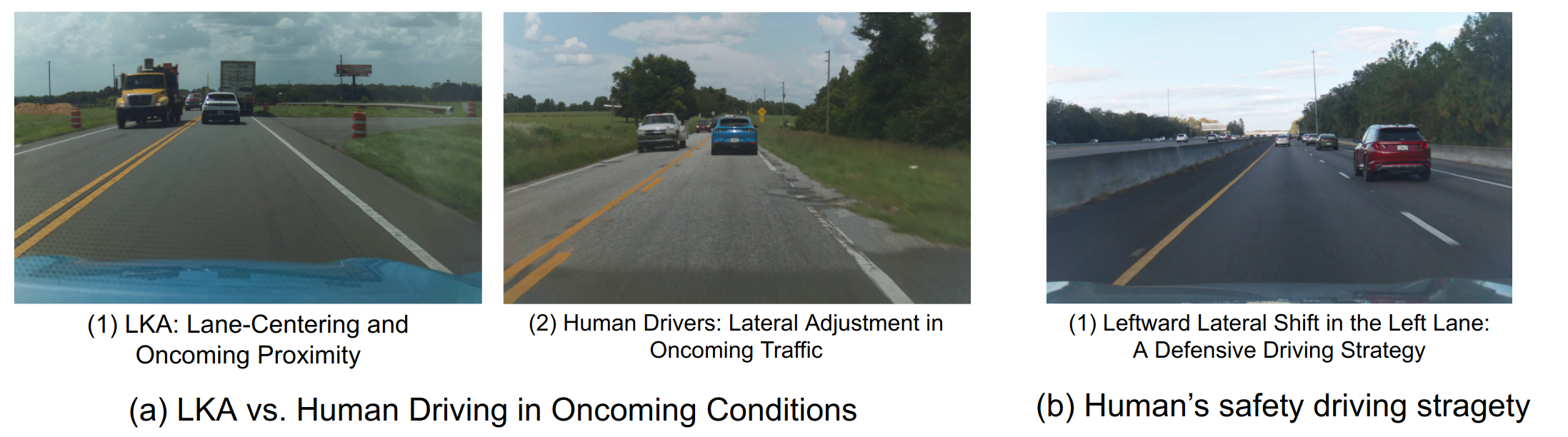}
  \caption{Comparisons of LKA and Human Driving's flexibility.}
  \label{fig:flexibleoncoming}
\end{figure}

Moreover, at on-coming traffic, avoiding oncoming vehicles and actively keeping a distance from side vehicles are also very important driving strategies. Figures ~\ref{fig:flexibleoncoming}(a) and (b) respectively show that LKA driving remains centered when the oncoming vehicle approaches the self-driving car, and that the human driver actively avoids the oncoming vehicle. Figure ~\ref{fig:flexibleoncoming} (c) shows the defensive driving strategy of humans compared to LKA. When the road space allows, they will drive to one side of the principle surrounding vehicles. 

It is worth mentioning that the current Tesla FSD system already has the ability to adjust steering angle on curves and actively avoid large vehicles or vulnerable groups around.

\begin{figure}[!h]
  \centering
  \includegraphics[width=0.48\textwidth]{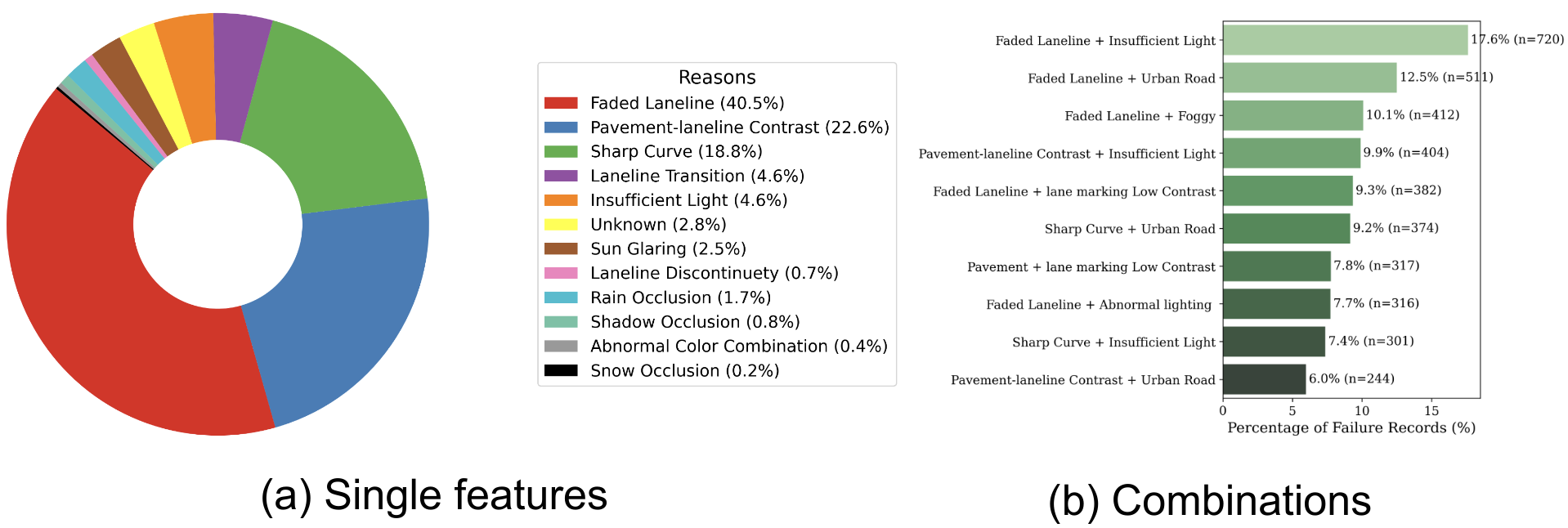}
  \caption{(a) LKA failure features in OpenLKA-Failure Dataset. (b) Most frequently occurring combinations of features in OpenLKA-Failure}
  \label{fig:reasons_distribution}
\end{figure}

\subsection*{Distribution of LKA Failure Conditions}
 
To understand the most common challenges faced by LKA in the real world, we perform statistical analysis on OpenLKA-Failure, illustrated in Fig.~\ref{fig:reasons_distribution} (a). The results reveal that LKA malfunctions under challenging conditions are predominantly attributed to several key factors: faded lane markings, low contrast between the pavement and lane markings, and sharp curves.
Furthermore, our analysis identified prevalent combinations of failure-inducing factors within OpenLKA-Failure, illustrated in Fig.~\ref{fig:reasons_distribution} (b).
Notably, the co-occurrence of faded lane markings with adverse environmental conditions---such as poor illumination, complex road geometries, or inclement weather---significantly exacerbates the system's susceptibility to failure.

\section{Discussions}

\subsection{Implications for Road Geometry Design and Speed Limits}
\label{subsec:geometry_design}


The lateral acceleration experienced by a vehicle depends on its speed $v(x)$, road curvature (via radius $R(x)$), and the roll angle $\text{roll}(x)$:

\begin{equation}
    a_{\text{lat}}(x) = \frac{v(x)^2}{R(x)} - g \cdot \text{roll}(x)
    \label{eq:acc_lat}
\end{equation}

Assuming that the steering torque $T(x)$ is proportional to lateral acceleration via gain $K_a$, we have:

\begin{equation}
    T(x) = K_a \left( \frac{v(x)^2}{R(x)} - g \cdot \text{roll}(x) \right)
    \label{eq:torque}
\end{equation}

Taking the spatial derivative with respect to $x$, and letting $R' = \frac{dR}{dx}$ and $\text{roll}' = \frac{d(\text{roll})}{dx}$:

\begin{equation}
    \frac{dT}{dx} = K_a \left( \frac{2 R v \frac{dv}{dx} - v^2 R'}{R^2} - g \cdot \text{roll}' \right)
    \label{eq:torque_dx}
\end{equation}

The temporal rate of change of steering torque is given by $\frac{dT}{dt} = v \cdot \frac{dT}{dx}$, where $a_x = v \cdot \frac{dv}{dx}$ is the longitudinal acceleration:

\begin{equation}
    \frac{dT}{dt} = v K_a \left( \frac{2 a_x}{R} - \frac{v^2 R'}{R^2} - g \cdot \text{roll}' \right)
    \label{eq:torque_dt}
\end{equation}

Under constant velocity ($a_x = 0$) and negligible roll gradient ($\text{roll}' = 0$), this simplifies to:

\begin{equation}
    \frac{dT}{dt} \approx K_a v^3 \left( \frac{-R'(x)}{R(x)^2} \right) = K_a v^3 \frac{d(1/R)}{dx}
    \label{eq:torque_dt_simple}
\end{equation}

\vspace{0.5em}
\noindent
\textbf{Design Remarks for LKA-Compatible Road Geometry.}  
Given that current LKA systems are primarily limited by their maximum steering angle and torque rate $\left(dT/dt\right)_{\max}$, we offer the following design-oriented recommendations:

\begin{enumerate}[label=\textbf{R-\arabic*:}, leftmargin=*, topsep=1pt]
    \item \textbf{Horizontal Radius ($R$):}  
    Governed by the steering angle limit. Road segments, especially at higher speeds, should use radii greater than the minimum threshold ($R_{\min} \propto v^2$) to provide adequate control margin for LKA systems.
    
    \item \textbf{Transition Curves ($L_s$):}  
    Governed by torque rate limits $(dT/dt)_{\max}$. Incorporate sufficiently long transition curves (e.g., spirals or clothoids) to ensure gradual curvature changes, minimizing the required torque rate. This is especially critical due to the $v^3$ dependence in Eq.~\eqref{eq:torque_dt_simple}.
    
    \item \textbf{Superelevation Development:}  
    Coordinate superelevation changes (road banking) smoothly with curvature transitions. Minimizing roll gradients ($\text{roll}'$) reduces lateral acceleration variability and the torque rate demands on the steering system, as reflected in Eqs.~\eqref{eq:torque} and \eqref{eq:torque_dt}.

    \item \textbf{Speed Limit Posting for Sharp Curves and  Poor Transitions:}  
    For curves where the radius $R$ approaches the lower bound of LKA-handlable curvature at typical speeds ($R_{\min}(v)$), consider posting lower advisory or mandatory speed limits to ensure feasible steering angles. In cases of short or missing transition curves—where $d(1/R)/dx$ is large—lower speed limits may be necessary to reduce the required $dT/dt$, which scales with $v^3$, and to maintain both LKA capability and driver comfort.
\end{enumerate}

\subsection{Distilling Empirical Observations into Predictive Insights: Lane Markings and Rural Road Safety}
\label{subsec:model_rural_safety}

While the primary focus of this paper is the empirical evaluation of LKA systems under real-world conditions, we take a further step to \textbf{distill these observations into actionable, generalizable insights}. To do so, we develop a machine learning–based model that captures the statistical relationship between observed external conditions and LKA performance outcomes, enabling broader application of our findings.

\begin{figure}[!h]
  \centering
  \includegraphics[width=0.48\textwidth]{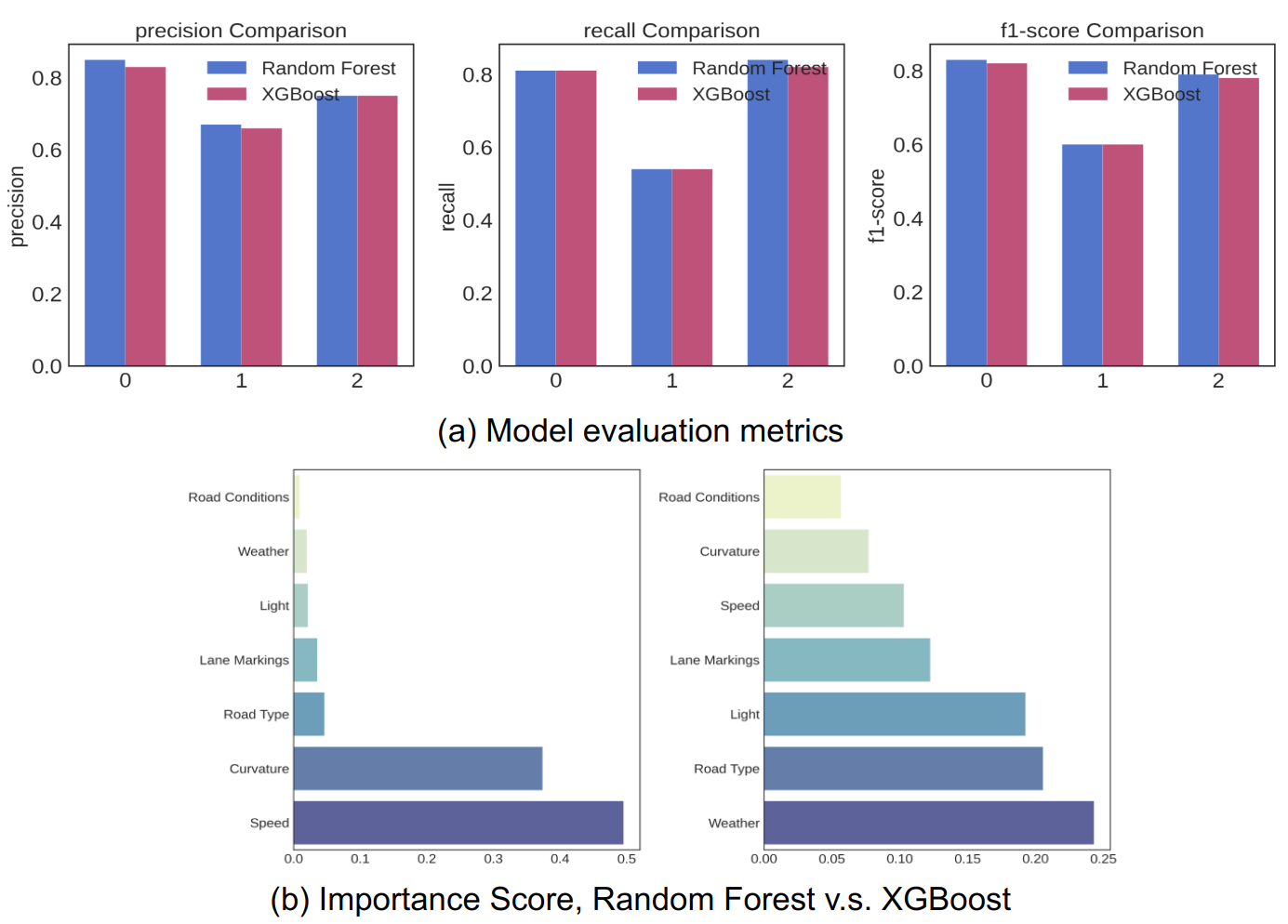}
  \caption{Performance Comparison of Random Forest and XGBoost Models for LKA Prediction}
  \label{fig:model}
\end{figure}

Using Random Forest and XGBoost classifiers, we train a predictive model on the OpenLKA-Failure dataset, where LKA performance is categorized into three classes: \textit{normal operation}, \textit{deviation} (lateral error $\geq$ 0.25 meters), and \textit{disengagement}. Input features include continuous variables (road curvature, vehicle speed) and categorical factors (road type, lane marking condition, lighting, weather, and surface quality).

\begin{figure}[!h]
  \centering
  \includegraphics[width=0.48\textwidth]{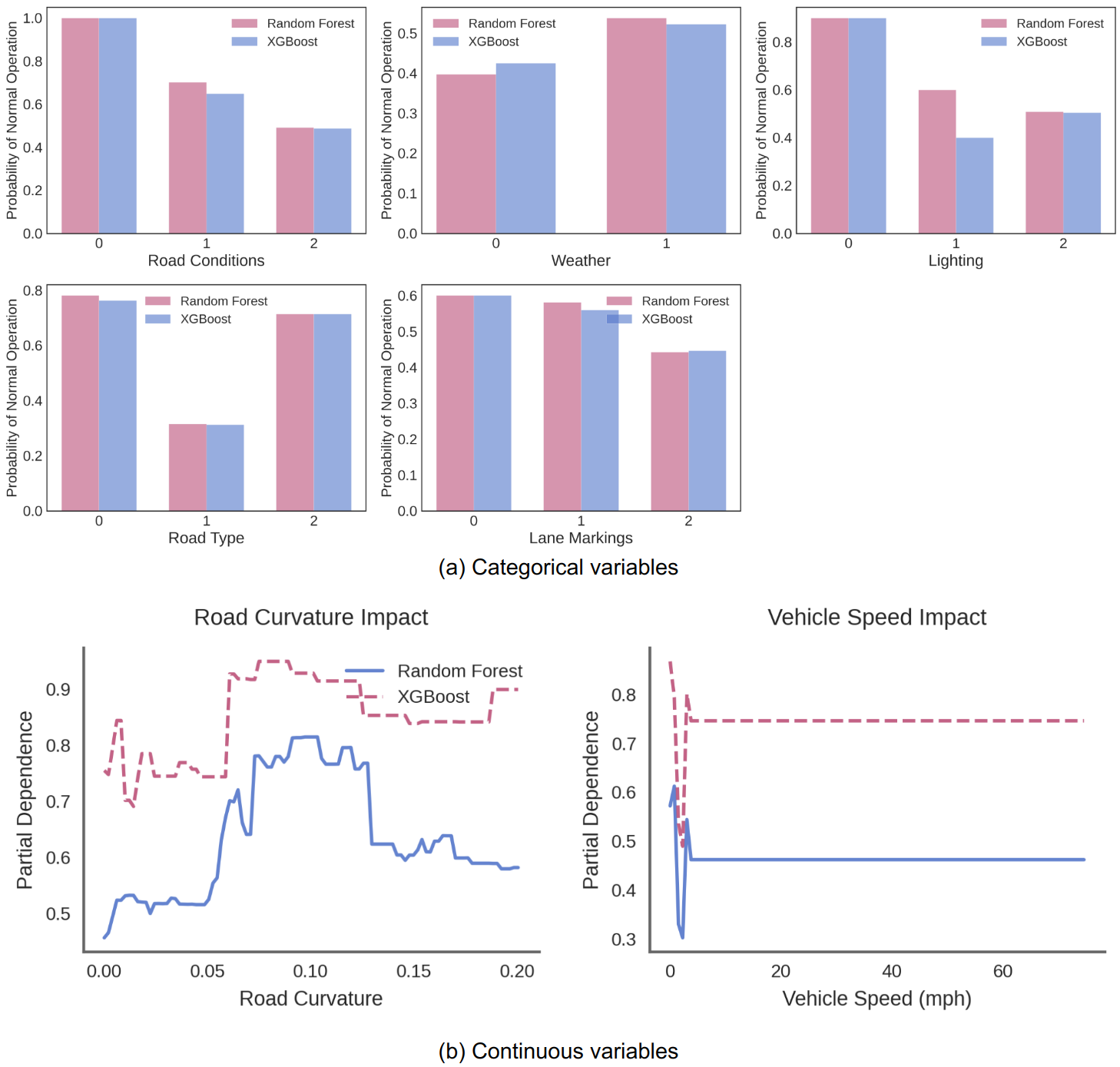}
  \caption{The impact of different variables on the model prediction results.}
  \label{fig:VImpact}
  \vspace{-0.5 em}            
\end{figure}

As shown in Figure~\ref{fig:model}, both models achieve high accuracy and recall in predicting performance categories. More importantly, variable importance and partial dependence analyses (Figure~\ref{fig:VImpact}) reflect our earlier empirical findings: road curvature and lane markings are dominant contributors to LKA failures, followed by weather and lighting conditions.

This predictive model serves as a bridge between empirical observation and proactive infrastructure planning. It enables agencies to assess LKA suitability on road segments where real-world testing may not yet exist. Based on model thresholds, we find that LKA deviation risk rises significantly when curvature exceeds 0.006~m$^{-1}$ and that reliable LKA operation typically degrades beyond 60.7~mph.

Finally, our findings have clear implications for rural road safety. Unlike urban roads, rural segments often lack consistent markings, lighting, or geometric regularity, factors that our empirical and model-based analyzes show are critical to LKA reliability. This highlights a growing concern about technological inequality: without targeted infrastructure improvements, rural areas may not fully benefit from ADAS capabilities. Our work provides both diagnostic tools and actionable insights to address this gap.

\section{Conclusion}
\label{sec:conclusion}

Our large-scale, real-world evaluation confirms that while modern LKA systems can contribute to safer driving, their reliability is highly dependent on roadway design, environmental and road conditions. In particular, faded or low-contrast lane markings, insufficient retro-reflectivity, and abrupt changes in road geometry frequently lead to lane deviations or system disengagements. Moreover, physical limitations in steering torque restrict LKA effectiveness on sharp curves, especially when transition lengths or superelevation development are inadequate. These findings highlight that \emph{road infrastructure is not just a boundary condition but a critical enabler, or limiter, of ADAS safety performance}.

From a transportation infrastructure perspective, our results suggest several concrete recommendations. \textbf{First, improve lane marking durability and visibility:} implement high-contrast, retro-reflective materials and prioritize consistent maintenance schedules, particularly on rural or lower-volume corridors where markings often degrade. \textbf{Second, refine geometric design standards:} ensure horizontal curvature, transition curves, and superelevation align with the operational limits of LKA systems, especially at higher design speeds. \textbf{Third, deploy intelligent speed warnings:} use curvature- and slope-aware advisory speed postings to alert both drivers and LKA-equipped vehicles of high-risk segments.

As a complementary direction, LKA system design must also evolve toward more adaptive control strategies, human-like trajectory anticipation, and multimodal perception augmented by infrastructure-integrated sensing (e.g., V2I beacons). Realizing the full potential of camera-based LKA systems will require coordinated progress in both infrastructure modernization and ADAS algorithm development.


\bibliographystyle{IEEEtran}
\bibliography{references}  

\end{document}